\pdfoutput=1
\documentclass[11pt]{article}

\usepackage{ACL2023}
\usepackage{times}
\usepackage{latexsym}

\usepackage[T1]{fontenc}
\usepackage[utf8]{inputenc}

\usepackage{microtype}

\usepackage{inconsolata}
\usepackage{enumitem}

\title{A Note on Bias to Complete}

\author{Jia Xu$^1$ \\
  Stevens Institute of Technology \\
  \texttt{jxu70@stevens.edu} \\\And
  Mona Diab $^2$ \\
  Carnegie Mellon University \\
  \texttt{mdiab@andrew.cmu.edu} \\}

\begin{document}
\maketitle

\begin{abstract}
Minimizing social bias strengthens societal bonds, promoting shared understanding and better decision-making. 
We revisit the definition of bias by discovering new bias types (e.g., societal status) in dynamic environments and describe them relative to context, such as culture, region, time, and personal background. 
Our framework includes eight hypotheses about bias and a minimizing bias strategy for each assumption as well as five methods as proposed solutions in LLM. 
The realization of the framework is yet to be completed. 
\end{abstract}

\section{Introduction}

\subsection{Main Objective} %0.33 page 
The bias that refers to an inclination or prejudice for or against one person or group is a major reason for societal division and injustice, as it affects our perceptions and decisions, often leading to unequal or unfair treatment of different groups and individuals. 
We aim to advance LLM technologies to alleviate communication barriers among different social opinions towards a connected community with two ultimate goals: (1) minimizing the disparity of recognizing social bias, contrasting  diverse perspectives on dynamically emerging bias types (often called sensitive or protected attributes); (2) eliminating the mis(dis)informed content verified from reliable sources. 

The notion of bias highly depends on the dynamic societal environments we live in, our ``context'', both geographically and temporally affected and impacted by events. For example, rights afforded to people of  different races change over time, and ethics standards and norms vary from place to place. 
%Typically such norms tend to be stable, however they can be modulated by episodic events and as well as the voices of influential entities.  
In addition to considering relatively well defined %We aim to steer away from an imposed and 
stationary value judgment of bias (e.g., gender and race) with stable and predetermined definitions (such as balanced ratios), we propose discovering new bias types ((e.g., poverty, educational level, societal status) and redefine them as conditioned on their context such as culture, region, time, and personal background. 
%and logical inferences reflecting realistic social norms. 

Around the central theme of relative bias in dynamic environments, we will provide a family of \textbf{8  hypotheses} forming our framework of communication and information processing for debiasing LLMs. To faciliate these hypotheses we will design novel NLP techniques for each of the hypotheses and deliver transformative methods and tools reflected in \textbf{5 methods} including bias definition, quantification, data augmentation, mitigation, reasoning, and analysis using mis(dis)information. The conceptual novelty lies in embracing the bias from \textbf{8 bias classes}: context, language, connotation, position, data, algorithmic, topic, and intention. 
In the following, we will first review the existing work limitation then highlight our innovations. 

\subsection{Current Limitations} %1 page
%\subsubsection{How is it done today, and what are the limitations?} 
\subsubsection{Key Technical Challenges}
%Each of our 6 tasks focuses on responding one limitation \textbf{(L)} as following:
We will focus on 5 aspects of limitations (\textbf{L}) of previous social bias work. %, and each of our task will respond to one limitation. 
%We address 3 limitations in current bias studies:
\begin{itemize}[leftmargin=*]
\vspace{-0.2cm}\item {\textbf{(L1)}}
Social bias has been manually pre-defined top-down and measured with a singular, fixed, and superficial social scale, almost devoid of context.  %\footnote{Richard Nordquist. 2019. Biased Language Definition and Examples. https://www.thoughtco.com/what-is-biased-language-1689168} % \cite{savoldi2021gender,stanovsky2019evaluating}. 
%First, existing work studies human pre-defined sensitive attributes, such as race and gender, while new attributes previously unconsidered may occur when the context changes,  such as poverty and club membership, which can result in biased views and discrimination. Moreover the interaction and intersectionality of such attributes have not been a focus of research to date.  
%Second, currently  
%and is mostly manually pre-defined top-down. 
However, in realistic settings, bias depends on, for example, cultural assumptions and inferences.  %(e.g., whose social role it is to propose in marriage) and varies over time (e.g., LGBTQ+ rights have drastically changed in the past 20 years). %all, please review whether it is appropriate as an example.
%Lastly, much-biased (whether positive or negative) information has been communicated through implication and connotation of utterances, necessitating the interpretation of communication to be within context. %ual interpretation particularly when put in the context. 
%on multifaceted social norms depending

\vspace{-0.2cm}\item {\textbf{(L2)}}
Measuring bias is challenging: first, quantifying bias lacks considering the social norm of the dynamic context; second, bias expressions can sometimes go through implicit language such as connotation; third, when the speaker, the listener, and the propagator are transferring  information, their different interpretations of words can cause biased viewpoints and misunderstanding. 

\vspace{-0.2cm}\item {\textbf{(L3)}}
Existing work in bias mitigation met three bottlenecks: first, the cause of social bias contains complex factors and has been understudied so far; second, artificially changing bias such as manipulating ratios across biased groups may cause injustice; third, sometimes bias is needed to adapt to a certain cultural background according to their social norm.

%For example, the cognitive bias, and the historical events such as colonialism, slavery having long-term effects on societal attitudes. Learning these factors will shut light on how to mitigate bias from the source. 

\vspace{-0.2cm}\item {\textbf{(L4)}}
While biased datasets are in high demand, obtaining labeled bias datasets can be complex due to privacy, cost, or legal considerations. Furthermore, existing data augmentation does not adapt to new bias types and definitions, social norms, as well as the dynamic content of the data, since the original frameworks are inherently static. %and the content of massive data from the web is constantly in flux. %The need to automatically generate attributes on the fly to discover previously undefined sensitive attributes is critical to building robust models. 

\vspace{-0.2cm}\item {\textbf{(L5)}}
The influence of mis/disinformation on bias and vice versa is understudied. Bias is often due to partial information, such as local news or minority viewpoints. 
Partial information or wrong information can lead to decision bias and discrimination. On the other hand, bias can sometimes cause misinformation or disinformation in communication,  %amplifying a misinterpreted connotation where it creates spurious correlations with potentially unwarranted  implied latent meaning.
whereas, the misinformation will likely travel around and even be included in a new LLM training worsening bias. 
\end{itemize}

\subsubsection{Proposed Solutions to Challenges} % 3.5 pages
%For each of the above challenges, we propose novel solutions for success, comprising 3 aims in our research plan. 
Each of our 5 methods focuses on responding one limitation \textbf{(L)} as follows:

% \begin{enumerate}
\begin{itemize}[leftmargin=*]
\vspace{-0.2cm}\item {\textbf{(M1)} {Contextualized bias definition with contrastive perspectives. }} 
We  veer away from the assumption that bias is absolute, instead developing a notion of relative bias against sets of norms of cultural standards and time. % all the while discovering emerging bias-sensitive attributes, such as combined attributes. %especially at the boundaries of the intersection of traditional well studied bias categories/attributes. 
We  will model a multi-faceted dynamic environment with relevant contexts (e.g., cultural, temporal, and regional) as a tensor as a means for grounding contrastive perspectives with a contextual backdrop. For example, an English statement is considered biased in a British Christian setting while the same statement might be considered appropriate within a British Muslim setting. 
Accordingly, We propose lifting the blindfold associated with LLMs by alleviating the potential blindspots inherently present in understanding generated text due to the biases in the minds of the reader or producer. %We will focus our exploration on written media. %, such as poverty in Babylon
% Western European normal within an Egyptian Middle Eastern Muslim setting. 
%Second, to find new bias terms, based on our prior work,\footnote{Wang, Sibo. Practice AI Responsibly with Proxy Variable Detection. BCG GAMMA 2021} we propose "proxy attributes via concept encoding" by devising latent encoded concepts from hidden features of pre-trained embeddings in transformers to discover new emerging sensitive attributes that if left undefined, could potentially lead to missing out on threats. Emerging sensitive attributes are a new form of bias in society that has not been discussed much previously. 

\vspace{-0.2cm}\item {\textbf{(M2)}{ Bias quantification with social norm and connotation understanding. }}
First, we will quantify the bias using multilingual consensus, implication understanding, geometric analysis on semantic and sentiment social norms.
%By considering and advancing algorithmic, language, connotation, position, date, topic, and intention bias, we will reduce the difference in social values with truth-supported norms for social harmony and justice. 
Second, we will take into consideration the speaker and the listener inherent perspectives as devices of interpretation and inference and factorize the relative bias to the derived connotations per utterance, namely devising a novel connotation-based tensor factorization model. 
Third, we will posit that a perceived bias in one context might be a norm in another based on the perspective such as personal background, and we will detect and fill in these blind spots to bridge the interpretations from the content reader and producer. %. For our technology to be resilient it needs to account for such contextualized bias
%, i.e. bias relative to a set of norms. 
%We will namely bring to the foreground the possible biases aassociated with the text shedding light on possible contextualized interpretations for the benefit of the reader. 
Theses methods will be realized for both individual and group as spheres of bias origin/impact and influence.

\vspace{-0.2cm}\item {\textbf{(M3)}{ Data generation using machine translation.}}
We will provide a novel approach to create labeled synthetic bias data by viewing debiasing as a multilingual machine translation task, considering the correspondence between the intersectionality of  contexts and language representation as the source while implications and connotations  rendered explicit are the target.
We will also infer the bias implicature by (machine) translating language into connotations and implications to explore the subtle types of implied language, incorporating secret languages or cryptolect.

\vspace{-0.2cm}\item {\textbf{(M4)}{ Bias mitigation and cause analysis. }}
First, we will classify the cause of bias using instructive reinforcement learning and provide explanations how bias gradually yields. 
Second, we will create a counter-intuitive method of removing bias-type input features on downstream tasks to minimize algorithmic bias. 
Third, we will implement a bias controller to tune on the perspective and adapt LLM output. 

\vspace{-0.2cm}\item {\textbf{(M5)}{ Mis(dis)information influence on bias.}}
%We will align the two to coordinate social opinions.  with the support of verified facts.
%We link such inferences to potential misinformation or malicious disinformation propagation. 
 %We will quantify and analyze the influence of such pathways in a manner akin to chain of thought and reasoning methods to control and model the integrity of the communication leveraging geometric methods modeling the information flow. 
%We propose an interpretable tensor-network model that will encode existing and discover new sensitive attributes relative to emerging dynamic social norms in LLMs. We will build contrastive and complementary models for both monolingual and multilingual LLMs.
Our model will control and guide information content along communication trajectories by understanding  how incomplete knowledge in users or LLMs could lead to unintended or misconstrued meanings of information, resulting in bias, and how bias can conversely aggravate mis/disinformation. 
\end{itemize}

\subsection{Novelty and Key Innovations} %1 page
Summarized, our project will lead to the following key innovations: 
\begin{itemize}[leftmargin=*]
\vspace{-0.2cm}\item { Novel concepts of contrastive bias definition and dynamic social context. }
\vspace{-0.2cm}\item { A theory of 8 hypotheses on LLMs and communication. }
\vspace{-0.2cm}\item { A paradigm by embracing 8 bias types for the issues arised in the hypotheses.} 
\vspace{-0.2cm}\item { New methods to realize our paradigm using tensor networks, machine translation, social norms, feature interpretation, connotation understanding,  mixture of experts, and so on. }
\vspace{-0.2cm}\item { Products for LLM debiasing either in 1 framework or separately used for 5 methods on bias definition, quantification, data augmentation, mitigation, explanation, and mis(dis)information. }
\end{itemize}

\section{Framework}

Our solutions are based on our framework  composed of lifting barrier framework and embracing bias strategy. 

\subsection{Bias Hypotheses}
Our framework comprises following hypotheses:

\begin{enumerate}[leftmargin=*]
\vspace{-0.2cm}\item {\textbf{Temporal Agnostic Context-Independence Assumption: }} 
LLMs often exhibit bias stemming from a lack of temporal context in processing prompts. Unlike humans, LLMs do not account for temporal gaps between inputs, potentially leading to erroneous context-dependent interpretations. 
The LLM's processing reflects a temporal context-independent bias, leading to potential misunderstandings or skewed context comprehension. 

\vspace{-0.2cm}\item {\textbf{Social Norm Alignment Assumption:}}
There exists a norm of social opinion on bias-related statements in a restricted environment. The median of embedding a statement over various biased statements with opinions can approximate this norm. The bias angle is the distance between the embedded biased statement and the norm. 

\vspace{-0.2cm}\item {\textbf{Individual Interpretation Assumption:}}
In connotation, the interpretation of a term by different people can vary, leading to a gap of bias and miscommunication. 

\vspace{-0.2cm}\item {\textbf{Information Over-writable Assumption:}}
A LLM can be fine-tuned and the new data will partially or even fully cover the data in the pre-trained model. Thus, information such as biased opinion can be revised or overwritten.  

\vspace{-0.2cm}\item {\textbf{Blindspot fill-in Assumption:}}
The blind spots on knowledge is one of the source of a biased view. Filling in the blind spots will shift the bias towards the norm.   

\vspace{-0.2cm}\item {\textbf{Bias No Effect on Performance Assumption:}}
In downstream applications, attributes like gender and race have minimal impact on the predictive performance of neural networks.

\vspace{-0.2cm}\item {\textbf{LowGram Phenomenon Assumption:}}
This term refers to a prevalent pattern in information dissemination, particularly within the domain of 'infotainment,' where content characterized by lower intellectual or cultural value (`light weight' and 'low taste') exhibits a higher propensity for rapid and widespread circulation in the public sphere. This phenomenon is particularly notable when such content requires minimal contextual support for its comprehension and propagation. The ``LowGram Effect" encapsulates the observation that simplistic, sensational, or superficial topics often achieve greater and faster distribution compared to more substantive, educational, or complex information, emphasizing the ease of dissemination of lower-quality, illusionary content over more intellectually rigorous material. 

\vspace{-0.2cm}\item {\textbf{Consistent Fact Narrative Validity Assumption:}} 
The validity of a narrative hinges on its construction from verified facts that demonstrate logical coherence. For a narrative to be considered valid, these facts must not only be consistent with each other but must also synergize to form a comprehensive and complete story. A narrative constructed from coherent facts but lacking in completeness may still hold validity, yet it remains unverified until it forms a fully realized and comprehensive story.
\end{enumerate}

We will verify these hypotheses on our dataset using deep learning and geometric methods on the embedding space. 

\subsection{Embracing Bias Strategy}

Our central theme is to embrace bias to mitigate bias. The rationale is that bias is mainly due to limited knowledge and narrowed view, and the inclusive of a broad spectrum of biased inputs will enrich the understanding of the ``world map" comprehensive to multi-angle perspectives, leading to a more objective conclusion on sensitive manners. We categorize the bias in the following types:

\begin{enumerate}[leftmargin=*]
\vspace{-0.2cm}\item {\textbf{Context bias}}
Bias based on context refers to the understanding and judgment influenced by the dynamic environment relative to the cultural, temporal, and regional information. 
We will collect data containing different dynamic environments to fill in our tensor factorization on the dependency modeling. The evaluation will also be carried out on multi-environments in multiple dimensions. 
\vspace{-0.2cm}\item {\textbf{Language bias}}
Bias from using different languages and perspectives arises because each language shapes thought differently, reflecting unique cultural contexts and values; for example, certain nuances or concepts may be inherent in one language but not easily translatable or understood in another. 
We will make use of the existing machine translation systems of co-PI Xu for multilingual LLM bias analysis and mitigation. 
The evaluation will be performed on English, Arabic, German, and Chinese on WMT datasets. 
\vspace{-0.2cm}\item {\textbf{Connotation bias}}
Bias from connotation and implication occurs when the emotional or cultural associations of words, beyond their literal meanings, influence perception and judgment, leading to interpretations that reflect personal or societal prejudices rather than objective assessments.
We will view the connotation as a machine translation task, where the source language is the connotation, and the target language is its debiased version in plain text. 
We will craft manual datasets on connotation examples for training and testing. The evaluation will measure the F-score on the label and prediction matches. 
\vspace{-0.2cm}\item {\textbf{Position bias}}
Bias from different personal experiences and viewpoints, combined with blindspots in knowledge, can significantly shape how individuals perceive, understand, and interpret information, where what is seen or understood is deeply influenced by what is known, believed, and overlooked.
We will measure the position of an individual or group view using the context-dependent bias definition to quantify the relative distance of a specific view to the social norm and evaluate manually with counterexamples.
\vspace{-0.2cm}\item {\textbf{Data bias}}
Bias due to data and domain occurs when the information used to train algorithms or make decisions reflects limitations, inaccuracies, or partial perspectives inherent in the data sources or the specific field of study, leading to skewed or prejudiced outcomes.
We will take data from multiple domains, such as news, TED talks, and scientific articles, to form multiple perspectives. Different experts and their backgrounds will agree upon manual evaluation. 
\vspace{-0.2cm}\item {\textbf{Algorithmic bias}}
Algorithmic bias refers to systematic and repeatable errors in a computer system that create unfair outcomes, such as privileging one arbitrary group of users over others. Here, we will consider the bias that comes from features and models other than data. 
We will eliminate algorithmic bias by excluding protected attributes such as gender and race when conducting predictions on downstream tasks of LLMs. Evaluation will be performed on criminal charge prediction and loan credibility.  
\vspace{-0.2cm}\item {\textbf{Topic bias}}
The "LowGram Phenomenon Assumption" affects the selection of topics in information dissemination by favoring unhealthy and unethical content that can spread quickly and widely, influencing public discourse and knowledge with less intellectually demanding and, figurative and vigorous content. 
We will classify the ethics of the topics and give preference to those that are more ethical and serious. 
We will construct sample tests on various topics and evaluate them manually. 
\vspace{-0.2cm}\item {\textbf{Blind spot bias}}
When given an utterance, we will decompose it into several statements with ideally small semantic and logic overlap. Then, for each statement, we verify its fact. We will form statement verification using a matrix, where the x-coordinate indicates each statement, and the y-coordinate indicates the viewpoints, such as different languages. The missing slot in the matrix is the blind spot, which will be filled by query knowledge using information retrieval, and the content of each of the slots will be checked against that of other slots with logical inference. Consensus slots will stay otherwise notified.
\end{enumerate}

\subsection{Related Work}

There have been extensive literature investigating bias issues in NLP, including but not limited to the following work:

\begin{itemize}
    \item Methods for objectively quantifying bias (e.g., relative to a specific collection of texts and other content): \cite{delobelle2021measuring,holtermann2022fair,weidinger2021ethical,fleisig2023fair,jourdan2023fairness,kong2023mitigating,krishna2022measuring,li2023prompt,teo2022fairness,zhao2023chbias,giguere2022fairness,kanekoc2022unmasking,tao2022powering,feng2022has,hu2022achieving,spliethover2022no,zhou2022towards,hana2021balancing,nejadgholi2022towards,stahl2022prefer,hanb2022fairlib,hessenthaler2022bridging,wang2021gender,du2021assessing,ahn2021mitigating,patel2021stated,spinde2022neural,zhong2021wikibias,levy2021collecting,wang2021eliminating,keidar2021towards,huang2021uncovering,he2021detect,shin2020neutralizing,chen2020detecting,manzoor2021uncovering,yuan2021simpson,pryzant2020automatically,dev2020measuring,yang2020causal,jiang2020reasoning,kim2022close,liu2021can,chowdhury2022learning,hiranandani2020fair,cho2020fair,mandal2020ensuring,ji2020can,ahmadian2020fair,ding2021retiring,singh2021fairness,esmaeili2021fair,zhang2021assessing,van2021decaf,hossain2021fair,maity2021does,meyer2021certifying,zuo2022counterfactual,ebadian2022sortition,wang2022usb,ray2022fairness,qi2022fairvfl,gardner2022subgroup,liu2022conformalized,psomas2022fair,lohaus2022two,ko2022all,gaucher2022price,soen2022fair,grazzi2022group,colombo2022best,chai2022self,xu2022deepmed,mehrotra2022fair,wang2022uncovering,an2022transferring,jesus2022turning,wang2022understanding,dooley2022robustness,van2022fair,czarnowska2021quantifying}
   %\cite{immer2021probing,immer2021probing,chalkidis2022fairlex}
    \item Computational techniques to characterize perspective spaces and measure the differences between perspectives: \cite{schrouff2022diagnosing,wan2021fairness,dullerud2022fairness,costa2022interpreting,ding2022word,xie2022fairness,cheng2022debiasing,wambsganss2022bias,liang2020monolingual,lei2022sentence,kanekob2022gender,chopra2020hindi,zhang2022effect,liu2022things,rungta2022geographic,blevins2022language,kim2022chapter,tzeng2020discovering,ma2021subgroup,chen2022fairness,gupta2022consistency,liang2022mind,mukherjee2022domain}
    \item Using human-LLMs interactions to identify analysts’ blind spots and induce perspectives representative of those blind spots: \cite{ma2023fairness,hertweck2022gradual,kanekoa2022debiasing,das2022quantifying,shwartz2020neural,wu2022context,lauscher2021sustainable,liu2021mitigating}
    \item Induction of outputs representing diverse perspectives (e.g., “How would a particular group or organization interpret this event?”): \cite{omrani2023social,feng2023pretraining,jiang2022generalized,elsafoury2022sos,kim2022close,he2022controlling,subramanian2021evaluating,zhang2020minimize,feldman2020designing,sen2020towards,amplayo2022attribute,bansal2022semantic,orgad2022gender,ziems2022moral,li2022keywords,zhao2022multihiertt,he2022bridging,pujari2022reinforcement,meissner2022debiasing,qian2022perturbation,gaci2022debiasing,he2022mabel,qiu2022late,ye2022zerogen,feng2022political,wu2022pcl,sridhar2022heterogeneous,wiher2022decoding,ruoss2020learning,el2020fairness,duan2021topicnet,venkatesh2021can,lapata2021structured,haghtalab2022demand,zou2022large,an2022cont,chen2022scalable,bommasani2022picking,mao2022enhance,chen2022pfl,gautam2022protovae,tan2022rise,espinosa2022concept,wu2022generative,chakraborty2022fair,aziz2022random,zhang2022fairness,jian2022non,zhang2022sampling,alghamdi2022beyond}
    \item Deriving insights from simulated dialogue between LLMs with different biases/perspectives: \cite{salewski2023context,hassan2021unpacking,weller2022pretrained,falk2022reports,rusert2022robustness,yeo2020defining,holtermann2022fair}
\end{itemize}

The existing work mostly focuses on the typical bias types, yet we are interested in new bias types and their flexible definition relative to the context. 

\section{Conclusion}

We introduce the notion of relative bias depending on the context, with thoughts on novel solutions to quantify and reduce LLM bias to join the effort for a more equal, fair, and connected society. 

\newpage
\bibliography{anthology,custom,literaturereview,references,biasllm}
\bibliographystyle{acl_natbib}

%\appendix

%\section{Example Appendix}
%\label{sec:appendix}

\end{document}